\documentclass[letterpaper, 10 pt, conference]{ieeeconf}  % Comment this line out if you need a4paper

\IEEEoverridecommandlockouts                              % This command is only needed if 
\usepackage[colorinlistoftodos]{todonotes}

\usepackage{caption}
\usepackage{subcaption}
\usepackage{amsmath}
\usepackage{amsfonts}
\usepackage{flushend}
\usepackage[hidelinks]{hyperref}
\usepackage[colorinlistoftodos]{todonotes}

\title{\LARGE \bf
DiPPeST: Diffusion-based Path Planner for Synthesizing Trajectories\\Applied on Quadruped Robots
}

\author{Maria Stamatopoulou$^{*}$, Jianwei Liu$^{*}$, and Dimitrios Kanoulas% <-this % stops a space
\thanks{The authors are with the Department of Computer Science, University College London, Gower Street, WC1E 6BT, London, UK. {\tt\small \{maria.stamatopoulou.21,jianwei.liu.21, d.kanoulas\}@ucl.ac.uk}}% <-this % stops a space
\thanks{$^{*}$equal contribution}
\thanks{This work was supported by the UKRI Future Leaders Fellowship [MR/V025333/1] (RoboHike) and the CDT for Foundational Artificial Intelligence [EP/S021566/1].  For the purpose of Open Access, the author has applied a CC BY public copyright license to any Author Accepted Manuscript version arising from this submission.}}

\begin{document}

\makeatletter
    \let\@oldmaketitle\@maketitle% Store \@maketitle
    \renewcommand{\@maketitle}{
    \@oldmaketitle
    \centering
    \vspace{-2.5pt}
        \includegraphics[height=2.52cm,width=2.89cm]{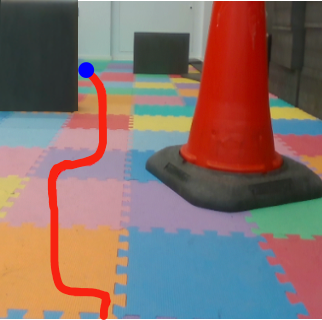}%
        \hfill
        \includegraphics[height=2.52cm,width=2.89cm]{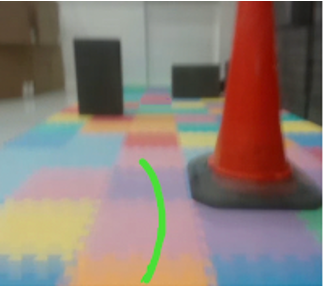}%
        \hfill
        \includegraphics[height=2.52cm,width=2.89cm]{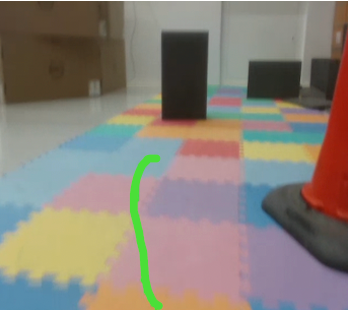}%
        \hfill
        \includegraphics[height=2.52cm,width=2.89cm]{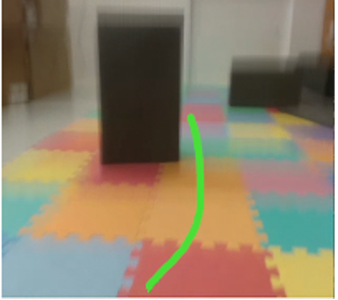}%
        \hfill
        \includegraphics[height=2.52cm,width=2.89cm]{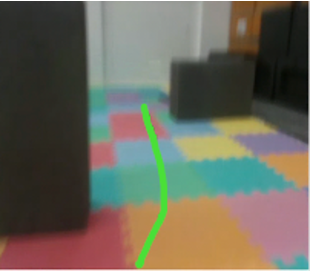}%
        \hfill
        \includegraphics[height=2.52cm,width=2.89cm]{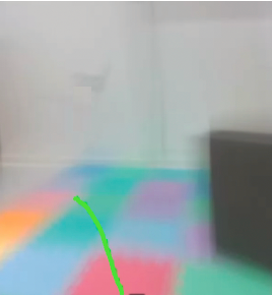}%
        \vspace{0.1cm}
        \includegraphics[height=2.8cm,width=2.89cm]{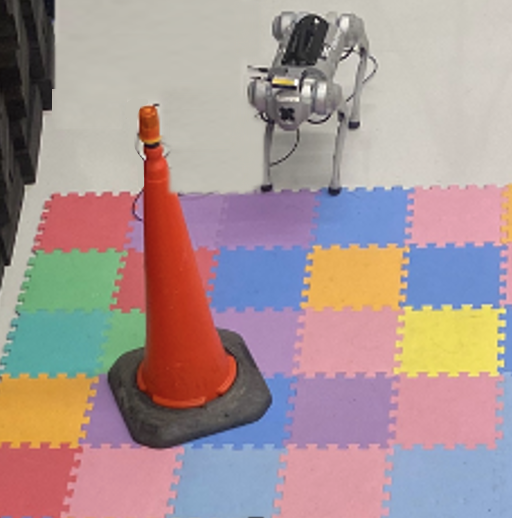}%
        \hfill
        \includegraphics[height=2.8cm,width=2.89cm]{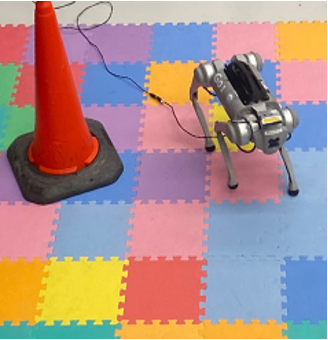}%
        \hfill
        \includegraphics[height=2.8cm,width=2.89cm]{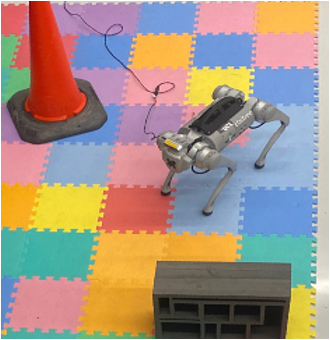}%
        \hfill
        \includegraphics[height=2.8cm,width=2.89cm]{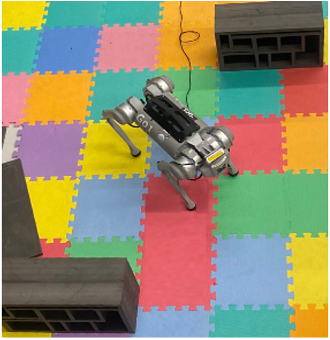}%
        \hfill
        \includegraphics[height=2.8cm,width=2.89cm]{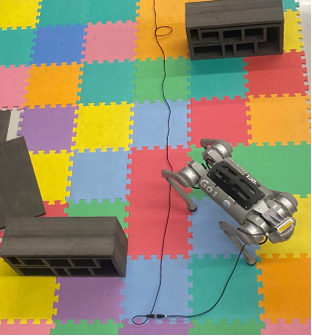}%
        \hfill
        \includegraphics[height=2.8cm,width=2.89cm]{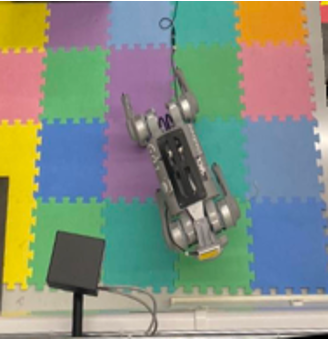}%
    \captionof{figure}{DiPPeST: (top) Consecutive robot camera frames with the overlaid generated global (red) and local (green) trajectories, captured in real-time; (bottom) A quadruped robot following the planned path. Blurriness is due to robot fast motion.}
    \label{fig:localpath}
    \vspace{-10.5pt}
    }
\makeatother

\maketitle
\thispagestyle{empty}
\pagestyle{empty}

%%%%%%%%%%%%%%%%%%%%%%%%%%%%%%%%%%%%%%%%%%%%%%%%%%%%%%%%%%%%%%%%%%%%%%%%%%%%%%%%
\begin{abstract}
We present DiPPeST, a novel image and goal conditioned diffusion-based trajectory generator for quadrupedal robot path planning. DiPPeST is a zero-shot adaptation of our previously introduced diffusion-based 2D global trajectory generator (DiPPeR).  The introduced system incorporates a novel strategy for local real-time path refinements, that is reactive to camera input, without requiring any further training, image processing, or environment interpretation techniques. DiPPeST achieves $92\%$ success rate in obstacle avoidance for nominal environments and an average of $88\%$ success rate when tested in environments that are up to $3.5$ times more complex in pixel variation than DiPPeR. A visual-servoing framework is developed to allow for real-world execution, tested on the quadruped robot, achieving $80\%$ success rate in different environments and showcasing improved behavior than complex state-of-the-art local planners, in narrow environments.
\end{abstract}

\section{INTRODUCTION}\label{sec:intro}
\setcounter{figure}{1}
Mobile robots have been gaining increased popularity due to their multi-purpose applications, varying from industrial production sites to search-and-rescue tasks~\cite{Kottege2022}. To successfully complete such tasks, real-time autonomous navigation in complex environments is required, rendering path planning an important open challenge~\cite{Cai2020}. Many works focus on generating global paths to provide a comprehensive path from a start to a goal point, assuming fully known and static environments~\cite{liu2024dipper, liu2023vit}  however, these methods cannot adapt in dynamic or unknown ones. Studies on local planners aim to address this limitation by utilizing real-time sensory data, enabling robots to make immediate decisions and adapt their paths in response to dynamic changes in their surroundings. However, these approaches may  sacrifice the optimality of the paths due to their myopic decision-making process. Hence, the ideal solution lies on the approach where the global planner sets an initial course path, and the local planner refines this path in real-time, ensuring both efficiency and adaptability. 
Learning from demonstration methods using image-conditioned Diffusion has demonstrated promising results in trajectory planning, primarily in the domain of manipulators~\cite{chi2023, carvalho2023, janner2022}, with few works extending these principles to mobile robot path planning~\cite{Sridhar_2023, liu2024dipper}. Diffusion policies iteratively calculate the action-score gradient based on visual observations, allowing for multi-modal action distribution representation, scalability to larger output spaces, and training stability while preserving expressive distribution capabilities~\cite{chi2023}. However, training a diffusion model is computationally expensive and timely, requiring large datasets, which makes their adoption more challenging than other methods.

In this paper, we present a novel approach to path planning by extending our global Diffusion-based 2D path planner, DiPPeR~\cite{liu2024dipper}), through a zero-shot adaptation that integrates an adaptive local path planner with no further training or fine-tuning. We introduce a robot path planning framework utilizing RGB data, complemented by a visual servoing pipeline for converting planned paths into actionable robot movements. We test our method on a real-world quadruped robot and compare the performance with state-of-the-art local planners both in static and dynamic environments. Our primary contributions include:
\begin{enumerate}
    \item A novel image-conditioned diffusion path planner for mobile robots for real-time path refinements.
    \item A zero-shot adaptation of DiPPeR to incorporate local path planning without re-training, fine-tuning, image pre-processing, or environment contextual or geometric information (e.g., semantics, obstacle recognition).
    \item A visual servoing real-world deployment stack.
\end{enumerate}

The remainder of the paper is structured as follows. In Sec.~\ref{sec:rw}, we briefly introduce literature in local path planning relevant to our proposed method. In Sec.~\ref{sec:pre}, we provide the necessary background knowledge, including a brief description of our previously introduced system, DiPPeR. In Sec.~\ref{sec:method}, we describe our proposed method, with our experimental results presented in Sec.~\ref{sec:exp}. Finally, in Sec.~\ref{sec:conclusion}, we summarize the results and conclude with some future work.
\section{RELATED WORK}\label{sec:rw}
Local path planning and visual servoing are extensively researched areas, as they enable mobile robots to make real-time decisions and adapt to dynamic environments efficiently. We briefly review these methods from both traditional and diffusion-based perspectives.

\subsection{Local Path Planning and Visual Servoing}
Traditional local path planning methods typically represent the environment using geometric primitives to identify traversal areas and obstacles~\cite{Kanoulas2019, Saeed_2020, Yang_2023}. However, these approaches often lack context awareness, characterized as the algorithm's inability to understand  the environment beyond physical obstacles and free spaces and assume a static environment. This leads to limited environment interpretation and exponentially increasing computational demands for more complex environments~\cite{Roth_2024}. 
One commonly used variation of geometric planners is the Dynamic Window Approach (DWA), where the robot's velocities are sampled to generate a trajectory optimized based on obstacle avoidance, reaching the target, and maintaining base velocity~\cite{Lee_2021, Missura_2019, Molinos_2019}. Chang et al.~\cite{Chang_2021} improved on the original DWA evaluation functions by employing Q-learning to adaptively learn the parameters, enhancing global navigation performance and efficiency. However, the method relies on the balance between effectiveness, speed, training quality and the complexity of the environment, which might limit its effectiveness in unpredictable or complex scenarios. More recent works attribute contextual information to gain a better understanding of the environment~\cite{Kantaros_2022}. Roth et al.~\cite{Roth_2024} introduced a learned local path planning approach, utilizing both geometric and semantic information to identify terrain traversability and obstacles effectively. Crespo et al.~\cite{Crespo_2020} proposed a path-planning approach employing co-safe Linear Temporal Logic over perception-based atomic predicates, enabling the generation of uncertain semantic maps for navigation. While semantic-driven sampling-based algorithms offer promising results, they heavily rely on the accuracy of the semantic map and the quality of sensor data, and they suffer from increased computational complexity under severe perceptual uncertainty. Our method aims to address the limitations of both semantic and geometric methods by not relying on any environment representation other than visual input and avoiding scaling computational complexity, as demonstrated in our previous work~\cite{liu2024dipper}.

Visual servoing provides a means of adaptively navigating based on visual cues~\cite{Johnson2018}. Traditional visual servoing techniques have been integrated with visual cognition towards path planning. For instance, closer to our method, Rodziewicz-Bielewicz et al.~\cite{Rodziewicz-Bielewicz_2023} utilize YOLO and real-time image processing from a central camera to determine the robot positions and orientations for precise trajectory planning and execution in indoor environments. However, the method is constrained by specific environmental conditions and the generalization capabilities of YOLO in complex environments. Zhu et al.~\cite{Zhu_2023} deploy a visibility-based search tree with a greedy search strategy that segments results based on field-of-view (FOV) constraints, ensuring the feature points remain within the camera's view. The approach assumes consistent visibility and the need to keep feature points within the FOV at all times, which may not be feasible in highly dynamic environments or in situations with occlusions, potentially leading to servo failures. Our methodology relaxes these dependencies, facilitating global and local path planning interaction. This ensures alignment with the overarching objective while simultaneously allowing for local refinements. 

\subsection{Diffusion for Path Planning}
Diffusion methods have gained popularity in the domain of path planning, with several works demonstrating promising results. Hong et al.~\cite{Hong2018} deploy diffusion maps to find local paths for reaching a goal while avoiding collisions with dynamic obstacles simultaneously, by computing transition probabilities between grid points. Liu et al.~\cite{liu2024dipper} introduce an image and goal-conditioned diffusion-based global path planner, trained using a CNN, showcasing great generalization and constant inference speed for various map types. Sridhar et al.~\cite{Sridhar_2023} present a unified robotic navigation diffusion policy that handles task-oriented navigation and task-agnostic exploration in unseen environments, exhibiting significant performance improvements and computational efficiency compared to existing approaches. We aim to leverage the advancement made in this diffusion-based path planning realm to develop our trajectory generator. While all these works require training of diffusion models, which can be computationally expensive and require large training datasets, our proposed approach does not require further training or fine-tuning of a diffusion policy. 
\section{Preliminaries}\label{sec:pre}

\begin{figure}[th]
    \centering
        \subfloat{
          \includegraphics[height=2.55cm]{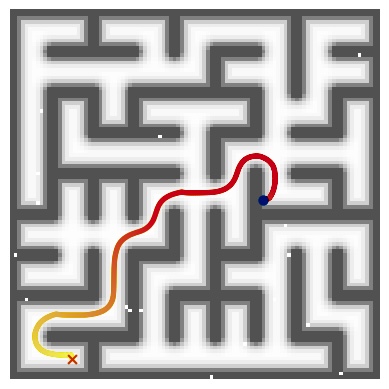}
        }
        \subfloat{
          \includegraphics[height=2.55cm]{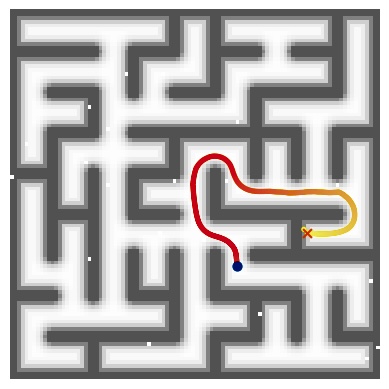}
        }
        \subfloat{
          \includegraphics[height=2.55cm]{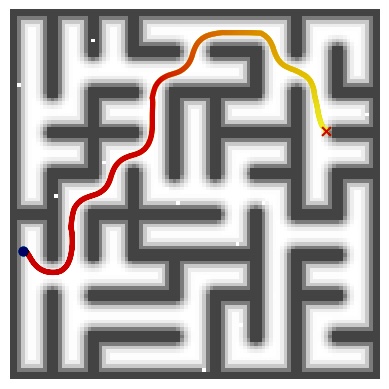}
        }
    \caption{Examples of DiPPeR's training dataset: $100\times100$ random solvable maps with examples of end-to-end trajectories, generated through $A^{*}$.}
    \label{Fig:dipper}
\end{figure}

Path planning involves computing a trajectory for a robot to follow, that is, optimizing for properties such as finding a collision-free, shortest, and smoothest route between start and goal positions. As elaborated in Sec.~\ref{sec:rw}, there are several approaches to solving path planning. For relevance to our method, we expand on our previous work in probabilistic diffusion-based path planning (DiPPeR~\cite{liu2024dipper}), which serves as a baseline for the developed DiPPeST system.

\begin{figure*}[ht!]
\includegraphics[width=1\textwidth,height=5cm]{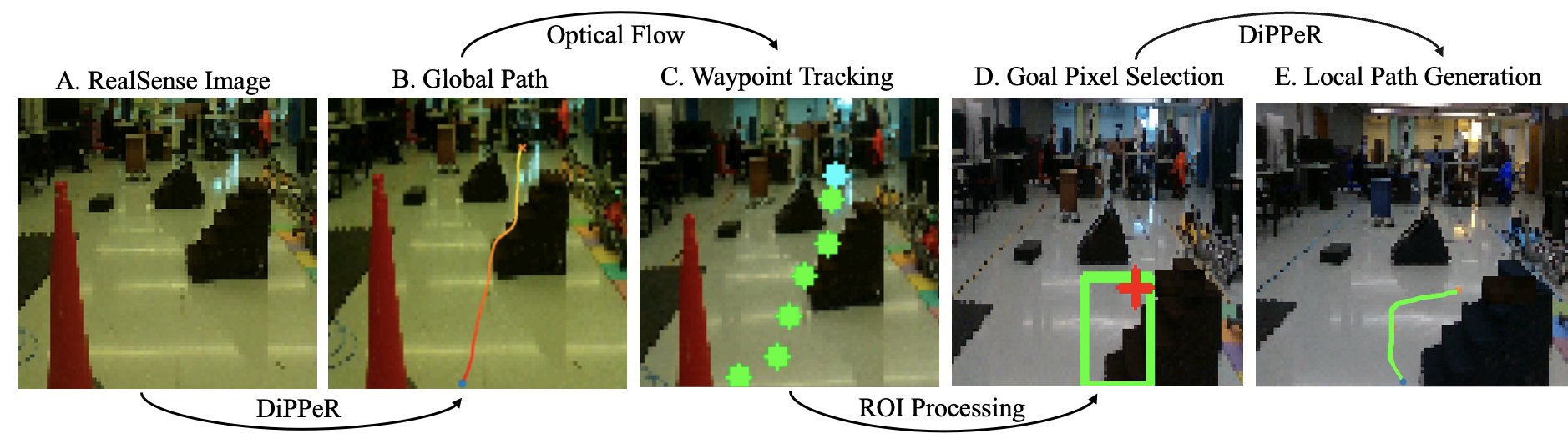}
    \caption{DiPPeST process of generating the local path and correcting global path failure. DiPPeST takes an input camera frame (A) and utilizes DiPPeR to generate a global path (B) and the waypoints. These are tracked at each input frame (C), and the optimal waypoint is selected. At each frame a, ROI is created around the current position and the waypoint and the optimal local goal position (D) within the ROI is selected (red cross) to avoid assigning the goal close to the obstacle due to (C) waypoint suboptimal positioning. DiPPeR is then used to generate a path to the goal position (E). Low resolution images are displayed representing the exact output of the diffusion model.}
    \label{fig:all_method}
\end{figure*}

DiPPeR is an image-based diffusion model designed to plan global paths from a starting to a goal position, given a 2D map in which space is classified as free or occupied. Trained on images of random mazes and ground truth trajectories generated by $A^{*}$ path planning, DiPPeR is able to generate paths via a Denoising Diffusion Probabilistic Model (DDPM)~\cite{hazart2023} -- a set of 2D path trajectories $A_{t}^{k}$ with added $\epsilon^{k}$ Gaussian noise and the map image observation $O$ are given as input, $K$ denoising iterations are performed via gradient descent, and a noise-free path trajectory $A_{t}^{0}$ is produced:
\begin{equation}\label{eqt:denoising}
    A_{t}^{k-1} = \alpha(A_{t}^{k}  - \gamma\epsilon_{\theta}(O,A_{t}^{k} ,k) + \mathcal{N}(0,\sigma^{2}I)),
\end{equation}
where $\epsilon_{\theta}$ is the noise prediction network (1D temporal CNN), $\alpha,\gamma,\sigma$ and the noise $\epsilon^{k}$ define the scheduling learning rate.

During training, Gaussian noise $\epsilon^{k}$ is iteratively added to the clean trajectory $A_{t}^{0}$. The noise predictor $\epsilon_{\theta}$ is trained using gradient decent of the mean squared error $\mathcal{L} = MSE(\epsilon^{k}, \epsilon_{\theta}(O,A_{t}^{0} +\epsilon^{k},k))$. DiPPeR achieves an average speedup of $23\times$ compared to state-of-the-art global path planning methods~\cite{liu2024dipper}. However, the method faces certain constraints. Initially, the path step number required for the denoising process must be empirically determined and manually selected for each experiment. This process is sub-optimal as it requires trial and error to determine the optimal length. Additionally, due to the inherent global planning nature of the approach, the goal position is manually chosen and remains unaltered throughout the experiment. This fixed setting proves suboptimal in scenarios requiring the dynamism of a local planner or in cases necessitating refinements.

% \subsection{Diffusion}
% Image-guided diffusion models~\cite{hazart2023} have emerged as a powerful generative model with impressive performance when dealing with image datasets, among others. They provide the ability to transform a latent encoded representation into a more meaningful description of the image data. A popular variation is the Denoising Diffusion Probabilistic Model (DDPM), a generative model defined through parameterized Markov chains trained using variations inference. A forward chain converts input data into noise and a reverse chain converts the noisy data back to its original form. In particular, the noisy data is generated by transforming the data distribution into a Gaussian distribution. Then, the denoising occurs by learning transition kernels parameterized using deep neural networks, for reversing the noisy data back to the input~\cite{yang2022}. The learned denoising kernel $p_{\theta}(x_{t-1}\mid x_{t})$ is parameterized by a prior Gaussian distribution $p(x_{T}) = \mathcal{N}(x_{T}:,0,I)$. Thus, the kernel can be defined by 
% \begin{equation}\label{diffusion_kernel}
%     \text p_{\theta}(x_{t-1}\mid x_{t}) = \mathcal{N}\left(x_{t-1} ; \mu_{\theta}(x_{t},t),\Sigma_{\theta}(x_t,t)\right)
% \end{equation}
% where, $\theta$ represents the model parameters, $\mu_{\theta}(x_{t},t)$ the mean and $\Sigma_{\theta}(x_t,t)$ the variance, parameterized by the deep neural network. A popular deep neural network choice for image conditioned diffusion are CNNs due to their benefits in dealing with image datasets.

\section{METHOD}\label{sec:method}
In this work, we utilize DiPPeR for global path and end-to-end trajectory generation while addressing its existing limitations. We further modify the framework to introduce DiPPeST, a diffusion-based local path planner for adaptive real-time local refinements leveraging visual input. Additionally, we incorporate a customized visual servoing strategy to ensure efficient execution of real-world paths. DiPPeST is a zero-shot adaptation of DiPPeR, requiring no further training or fine-turning while introducing a novel local planning framework.

\subsection{Global Planner}~\label{subsec:globalpathgeneration}
Firstly, our method involves the generation of a global path plan based on RGB input camera data. This path is generated via DiPPeR, utilizing the first captured frame from the robot's onboard camera which is positioned at it's front, facing towards the goal. The model generates a feasible path that avoids all obstacles while aiming to reach the set goal, by maintaining local uniformity in trajectory generation and traversing through pixels with consistent intensity levels. Unlike DiPPeR's original method, a map of the environment is not required, and planning happens directly through the RGB frame of the camera. Despite significant variations in the input images, DiPPeR demonstrates zero-shot adaptation to the robot's perspective and RGB image inputs despite being initially only trained on top-down grayscale maze images (Fig.~\ref{Fig:dipper}). This transferability likely stems from the models' property to prioritize the generated trajectory's consistency and goal-reaching condition and its ability to learn color gradient-based features through ResNet, which are transferable from grayscale to RBG. Further exploration of the model's generalization capabilities is provided in Sec.~\ref{subsec:genaralisation} (also demonstrated in Fig.~\ref{fig:globalpath}). 

\subsection{Local Planner}
Given the generated global path, frame-to-frame local path adjustments are required to better navigate around obstacles and adapt to dynamic environments as the robot progresses toward its goal. To address this, we introduce DiPPeST, which takes an RGB image as input and outputs a 2D local goal position along with a 2D feasible collision-free trajectory to be followed, all within the camera frame. The method involves tracking the global path waypoints and the global goal position within individual local camera frames. At each frame, a strategy is devised to select the optimal waypoint, and a region of interest (ROI) is created from the robot's current position to the selected waypoint position. Processing within this ROI determines the optimal local goal position for the robot to reach in the current frame and DiPPeR generates the feasible path to the goal. An overview of DiPPeST is illustrated in Fig.~\ref{fig:all_method}.

\subsubsection{Waypoint Tracking}\label{Sec:waypoints}
We utilize the Lucas-Kanade optical flow estimation with the pyramidal approach for tracking the global path features across frames~\cite{Lucas-Kanade}. The method assumes that optical flow remains constant within a local neighborhood of pixels. Let $I_{x}$ and $I_{y}$ denote the gradients of the image in the x and y directions, respectively, and let $I_{t}$ represent the temporal gradient between consecutive frames. The linear optical flow vector $(u,v)$ is given by: 
\begin{equation}\label{eqt:opticalflow}
    I_x \cdot u + I_y \cdot v = -I_t
\end{equation}
This can be estimated by solving the system of linear equations for each pixel via the pyramidal approach, i.e., building an image pyramid, where each level is a down-sampled version of the previous one. The optical flow is estimated at each level iteratively. The waypoints are then tracked by individual frames as seen in Fig.~\ref{fig:all_method}-C.

\subsubsection{Intermediate Waypoint Selection}~\label{subsec:waypointselection}
At each frame, the optimal waypoint to follow is determined based on two conditions: (a) the half-point distance $dist$ from the current position to the furthest away waypoint and (b) the highest similarity $sim$ to the direction vector $\textbf{dir}$ of the goal position. This ensures that for each frame, the planned trajectory will aim toward the goal position while allowing for a significant look-ahead distance for refinements.  
For condition (a) the Euclidean distance $dist$ is calculated between the current position $\textbf{p}_{curr}$ and all $n$ waypoint positions $\textbf{p}_{way}$ as:
\begin{equation}\label{eqt:distance}
 dist = \sqrt{\sum_{i=1}^{n} (\textbf{p}_{curr} - \textbf{p}_{way}[i])^2}.
\end{equation}
The median distance is then selected as the optimal waypoint position. Choosing the median distance waypoint helps avoid myopic local decisions that may be sub-optimal on a global scale. 
For condition (b), the dot product between the current position $\textbf{p}_{curr}$ and all $n$ waypoint positions $\textbf{p}_{way}$ is calculated as follows: 
\begin{align}\label{eqt:direction}
    & \textbf{dir} = \textbf{p}_{curr} -\textbf{p}_{way} \\
    & \textbf{p}_{norm} = \frac{\textbf{dir}}{\| \text{\textbf{dir}} \|_2}, ~~~~ \textbf{g}_{norm} = \frac{\textbf{g}}{\| \textbf{g} \|_2} \\
    & sim = \textbf{p}_{norm} \cdot \textbf{g}_{norm}^T
\end{align}
where $\textbf{p}_{norm}$ is the norm direction, $\textbf{g}$ represents the goal position, $\textbf{g}_{norm}$ is the norm goal, and $\cdot$ the dot product. Finally, the next waypoint $i$ ($\textbf{p}_{way}[i]$) with the median distance $dist$ and highest similarity $sim$ is chosen as the next intermediate goal.

\subsubsection{ROI Processing and Goal Selection}~\label{subsec:roiinterpretation}
To ensure that the local goal is set optimally within a traversable region and to account for the possibility of a waypoint being incorrectly positioned within an obstacle, we further condition its selection. A ROI is defined in the camera frame, spanning between the current robot position and the selected intermediate waypoint. This ROI segment is utilized to assess the variation in pixel intensities. Let $px_{std}$ be the standard deviation of pixel values in each color channel:
\begin{equation}
    px_{std} = \sqrt{\frac{1}{N} \sum_{i=1}^{N} ({ROI}_i - \overline{ROI})^2}
\end{equation}
where $N$ denotes the number of pixels, ${ROI}_i$ is the pixel value at position $i$ and $\overline{ROI}$ is the mean pixel value in each color channel. The normalized standard deviation ($px_{var}$) is then obtained by dividing $px_{std}$ by the maximum possible pixel value, typically $255$ for $8$-bit images. This serves as a quantitative measure of intensity variation within the ROI. The value of $px_{var}$ is used in Sec.~\ref{subsec:localpathgeneration} to determine the number of path steps during the denoising process.

To determine the local optimal goal position, the most common pixel intensity $px_{int}$ within the ROI is obtained by calculating the mean pixel intensity $\overline{px}$ within the ROI. To maintain consistency across frames, the previously selected pixel intensity $px_{pint}$ serves as a prior and is used as a tolerance value $\tau$ within which the new $px_{int}$ should be set. 
\begin{equation}
   px_{\text{int}} = \text{mode} \left( \{ px \mid px \in ROI, |\overline{px} - px_{\text{pint}}| \leq \tau \} \right)
\end{equation}
This prevents setting the local goal within an obstacle in a ROI that is predominantly surrounded by obstacles while also maintaining local consistency for the diffusion model and allowing the flexibility to navigate in traversable regions of varying intensities. The optimal pixel $px^{*}$ needs to have intensity $px_{int}$ while also being the closest to the waypoint position, ensuring that the selected goal does not deviate significantly from the global path. The $px^{*}$ is chosen as the next local goal, and its coordinates are scaled back to the entire image frame. An example of this correction is visualized by the red cross in Fig.~\ref{fig:all_method}-D.

\subsubsection{Local Path Generation}~\label{subsec:localpathgeneration}
After the new goal position is selected, DiPPeR is utilized to generate the feasible path, involving path step selection and path correction. Firstly, the path step number $ps$ is selected through a combination of the Euclidean distance between the current position to the local goal position, $px_d$ and the $px_{var}$ within the ROI following: 
    \begin{equation}
    ps = px_d + \alpha e^{(10 \times px_{var})}
\end{equation}
 This method addresses the manual path step number selection of DiPPeR and allows it to increase exponentially as the variation of pixel intensity increases, thereby providing adaptation space scalable to obstacle density. We set  $\alpha$, the scaling coefficient, equal to $4$, to agree with the input dimensions of the network. In cases where the local trajectory fails to avoid obstacles, a mitigating method is implemented to allow for re-planning, ensuring that the robot follows only feasible trajectories. The generated local trajectory is evaluated within the ROI to ensure that it does not traverse pixels whose intensity falls outside of the tolerance range compared to the local goal position pixel intensity. 

\subsection{Path Following}
To execute the local paths generated by DiPPeST in the real world, the trajectory coordinates need to be de-projected from 2D image coordinates $(x,y)$ obtained by the camera mounted on the robot's head into 3D world coordinates.
\begin{figure}
\centering
\includegraphics[height=9cm]{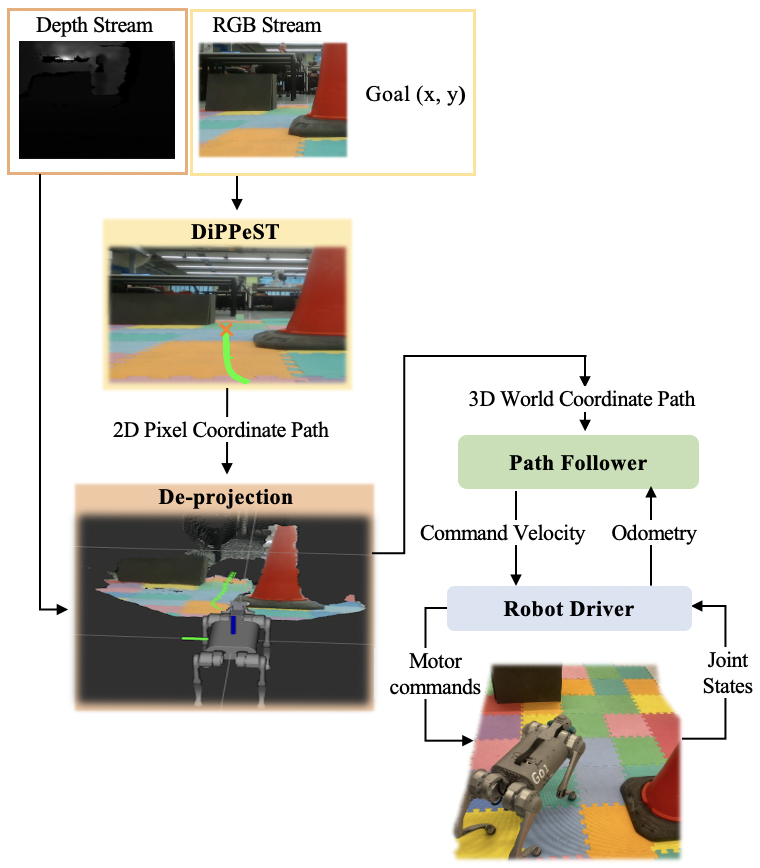}
\caption{Visual servoing framework for real-world robot path execution. DiPPeST 2D trajectories within RGB frame are de-projected into 3D world coordinates. The path follower translates the trajectories into velocity commands, and the robot module into motor commands to drive the robot towards the desired path.\label{fig:servoing} }
\end{figure}
Given the depth $d$ from the depth reading for pixel coordinate $(i,j)$ of the camera:
\begin{equation}
\text{Deproj}(i,j,d) = \left( d \cdot U_{\text{Model}} \left( \frac{(i,j) - P}{F} \right), d \right)
\end{equation}
where $P =(p_{x}, p_{y})$ is the principal point $F = (f_{x}, f_{y})$ the focal length, both intrinsic parameters of the camera, $U_{\text{Model}}$ represents the model for lens distortion, and $d$ is the depth obtained directly from the depth image. 
The resulting de-projected 3D local path is then executed by a path follower module~\cite{Yang_2023}, which generates velocity commands from the input path. These commands utilize odometry readings to adjust the robot's movement in real time, ensuring accurate adherence to the planned trajectory.
Finally, the robot driver module translates the velocity commands into motor commands, considering the robot's joint states to drive the hardware toward the desired state as defined by the 3D trajectory. A twist-command correction is implemented to ensure the robot's view does not significantly divert from the global goal position. An overview of the framework is depicted in Fig.~\ref{fig:servoing}.

\section{RESULTS}\label{sec:exp}
\subsection{Performance Evaluation}\label{subsec:genaralisation}
\begin{figure}[th]
    \centering
       \subfloat[Variation of Input Image PoV]{%
          \includegraphics[height=2.6cm,width=2.8cm]{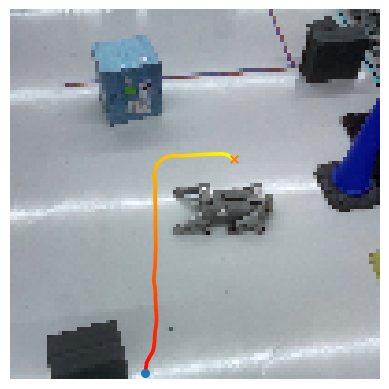}%
          \includegraphics[height=2.6cm,width=2.8cm]{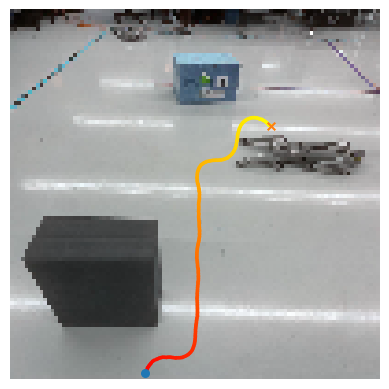}%
          \includegraphics[height=2.6cm,width=2.8cm]{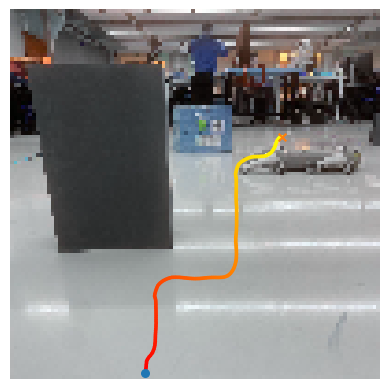}%
          \label{subfig:localpov}%
       }
       \quad
       \subfloat[Variation of Input Image Size]{%
          \includegraphics[height=2.6cm,width=2.8cm]{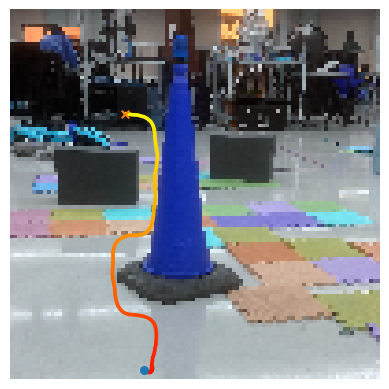}%
          \includegraphics[height=2.6cm,width=2.8cm]{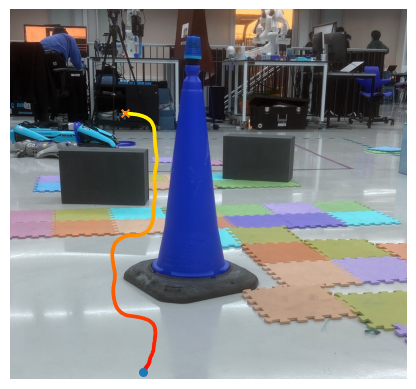}%
          \includegraphics[height=2.6cm,width=2.8cm]{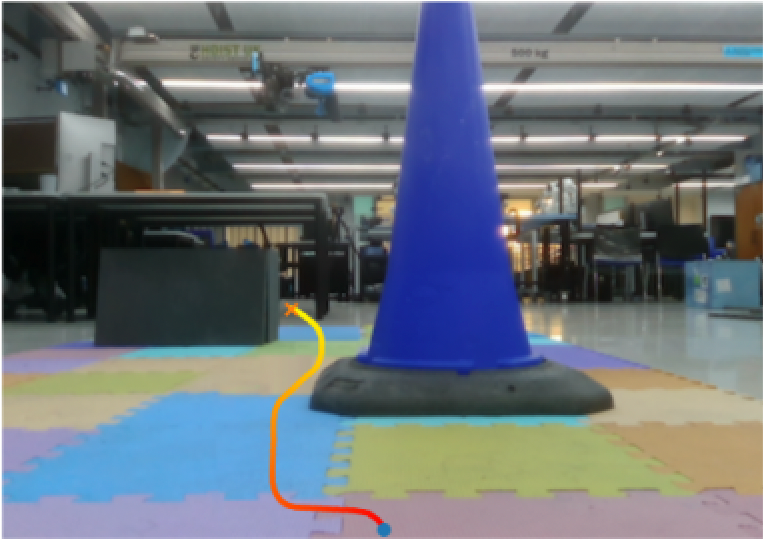}%
          \label{subfig:localsize}%
       } 
        \quad
        \subfloat[Edge Cases]{%
          \includegraphics[height=2.6cm,width=2.8cm]{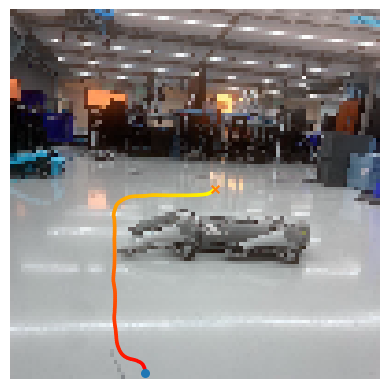}%
          \includegraphics[height=2.6cm,width=2.8cm]{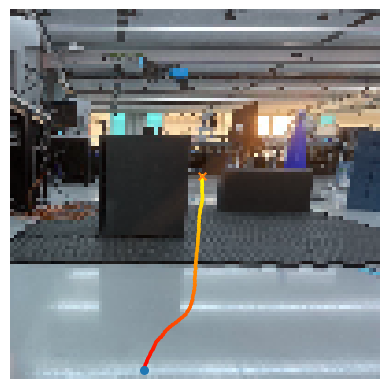}%
          \includegraphics[height=2.6cm,width=2.8cm]{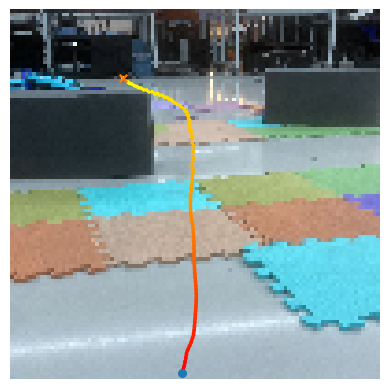}%
          \label{subfig:locaedge}%
        }
    \caption{Illustration of DiPPeST generated trajectories for cases of variation of a) of camera PoV: top-down, human-eye, robot-eye, b) input image size: training dataset, iPhone 11, RealSense D435i, and c) edge cases. Low resolution frames correspond to the compressed images as received for the denoising process.
    \label{fig:edge_cases}}
    \label{fig:globalpath}
\end{figure}

We conduct experiments to assess the robustness of DiPPeST against varying input conditions and validate its performance and generalization capabilities in out-of-distribution cases. These experiments are performed using the RGB frames obtained through a RealSense D435i camera. We focus on a) variation of the traversable region and obstacle pixel intensity, b) change of input image size, and c) variation of camera point-of-view (PoV), as seen in Fig.\ref{fig:edge_cases}. We quantify the effect as the average percentage of successful collision avoidance attempts over $10$ trials for each case. DiPPeST is executed on a computer with an Nvidia RTX 3090 GPU and has a constant inference time of $0.42$s per frame, as it depends on DiPPeR's denoising time of $0.4$s.

We examine the impact of varying the color of the traversable regions and obstacles in DiPPeST trajectory generation performance, as DiPPeR was only trained on black and white images (Fig.~\ref{Fig:dipper}), with white representing the feasible path and black the obstacles. The color of the traversable region is varied by adding multi-color pads, introducing varying pixel intensities within the ROI. The mean variance values are converted into percentage values to standardize comparisons. DiPPeST's performance is quantified by assessing the correlation between changing the percentage ROI pixel variance and the success rate. DiPPeST achieves a mean success rate of $85\%$ for generating successful trajectories, showing good generalization capabilities when considering the training dataset. It successfully avoids obstacles with a $92\%$ success rate for an ROI variance of less than $30\%$, representing the most common environmental scenario. Increasing pixel intensity variation results in a $19\%$ decline for the edge case of ROI variance exceeding $80\%$, indicating a slight negative trend. This trend could potentially be corrected by introducing colorful maps into the training dataset. 

\begin{figure}[th]
\centering
\includegraphics[height=5.5cm,width=7cm]{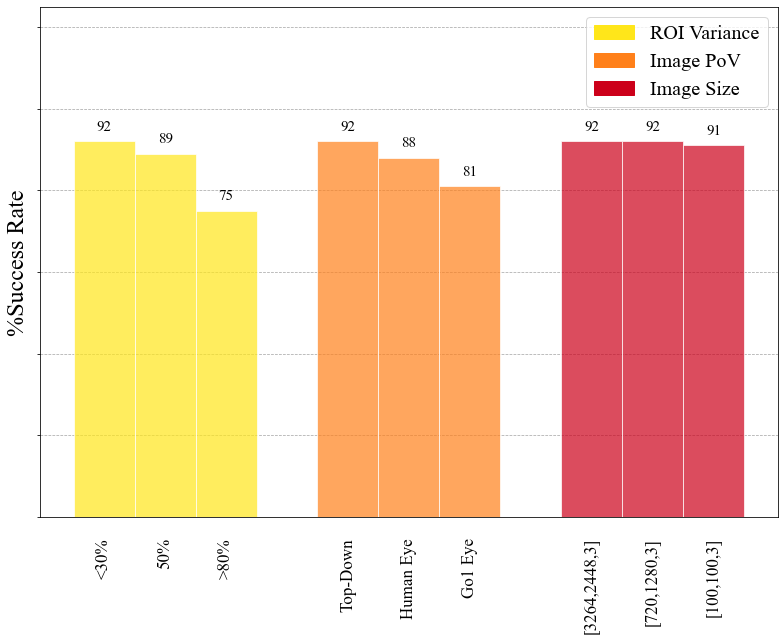}
\caption{The effect of variation of a)  floor and obstacle color, b) input image size, and c) camera PoV, over DiPPeST $\%$ success rate. \label{fig:ablation}}
\end{figure}

We further evaluate the impact of image PoV on the success rate, considering that DiPPeR is trained on maps with a top-down PoV. DiPPeST should generalize to input images of varying PoV, reflecting variations in camera angle and heights in real-world scenarios. To assess this, we conduct the experiments on recorded RGB frames obtained by changing the angle and height of the position of the camera, as depicted in Fig.\ref{subfig:localpov}. DiPPeST generates successful trajectories with an average $87\%$ success rate for all PoV variation experiments. There is a slight negative trend as the PoV angle changes from top-down to robot-level with an overall $13\%$ decline in success rate, which could be corrected by refining the training dataset.

We also validate DiPPeST's performance for image inputs of variable sizes, as DiPPeR is trained on maps of the same size. DiPPeST receives camera input, which may have a variable FoV based on hardware specifications. This concern is based on the compression ratio variations occurring when the image input is passed on the encoder network. We examine images of varying sizes, including those of the same size as the training data set $[100,100,3]$, images taken from an iPhone 11 camera $[3264, 2448, 3]$ and a RealSense camera $[720, 1280, 3]$, as seen in Fig.\ref{subfig:localsize}. However, there is no significant effect on the success rate, with all scenarios displaying a $92\%$ generalization success rate. 

Overall, DiPPeST achieved an average success rate of $92\%$ for standard environments, surpassing DiPPeR's validation performance. Additionally, it achieves an overall generalization success rate of $88\%$. These results are summarized in Fig.~\ref{fig:ablation}.

We also investigate the edge cases where the obstacles are similar in pixel intensity to the traversable region. DiPPeST's trajectory generation capabilities for these cases are shown in Fig.~\ref{subfig:locaedge}. In the first subfigure of Fig.~\ref{subfig:locaedge}, the obstacle and the floor have an average pixel intensity difference of $26\%$ for all three channels, and on the second subfigure, a difference of $28\%$. Given that the intensity difference of DiPPeR validation maps is $97\%$,  DiPPeST showcases a $73\%$ superiority in generalization capabilities with an improvement factor of $3.5$. 

\subsection{Real-World 
Evaluation}\label{subsec:inference}
For real-world evaluation, the Unitree Go1 robot is used with DiPPeST, taking as input images from an Intel RealSense D435i camera mounted on the front at an angle of $10$ degree depression. For all experiments, the global plan is generated from the first frame (Sec.~\ref{subsec:globalpathgeneration}) while the robot remains stationary. 
% Once the robot starts moving, the local planner starts generating the next local goal position and trajectory while aiming to reach the goal position (Sec.~\ref{subsec:localpathgeneration}).
We test DiPPeST's performance for a) static environments and b) dynamic environments by measuring the rate of successful attempts in avoiding obstacles and reaching the goal position across all attempts.
For static environment experiments, we test our method in lab conditions by constructing a course with stationary obstacles. For dynamic environments, walking robots are added within the scene to introduce dynamic obstacles. Each experiment is performed for $3$ different instances of each environment, with increasing obstacle density and decreasing obstacle distance to showcase avoidance in narrow environments. The results are presented in Table~\ref{tab:sim_run_time_comparision}.

\begin{table}[hbt!]
    \centering
    \begin{tabular}{ |c| c c c c|c|} 
     \hline
     Environment & sCA & sGR & dCA & dGR & \%mSR \\ [0.5ex] 
     \hline
     DiPPeST & \textbf{85}  & 79  & 78  & \textbf{79} & \textbf{80.3} \\ 
     IPlanner & 76   & \textbf{82} & 67   & 78  & 75.6 \\
     NoMad & 83  & 71  & \textbf{83}  & 73 & 77.5 \\
     \hline
    \end{tabular}
    \caption{This table presents DiPPeST real-world Collision Avoidance (CA) and Goal Reaching (GR) mean \% success rates (\%mSR), compared to other SOTA methods. \label{tab:sim_run_time_comparision}}
\end{table}

DiPPeST achieves an overall success rate of \textbf{80\%} for all environments, showing improved behavior against IPlanner, a geometric-based local planner and  NoMad, a diffusion-based local planner. Both methods require training for optimal planning as opposed to DiPPeST. IPlanner displayed capabilities in reaching the goal, however it often stumbled upon obstacles while doing so. On the other hand, NoMaD demonstrates successful obstacle avoidance in both dynamic and static environments, but it would get stuck when reaching too close to a non-traversable area. Both SOTA methods avoid the narrow passage and redirect following a safer and less efficient path. An example of an executed DiPPeST path is depicted in Fig.\ref{fig:localpath}. DiPPeST's failure case in goal reaching is attributed to the requirement that the global goal should not significantly deviate from the view of each image frame, as well as restrictions on re-planning speed, which might be limiting if the speed of the dynamic obstacles is high.

\section{CONCLUSION}\label{sec:conclusion}
In this work, we introduced DiPPeST, a novel image-guided diffusion-based method for synthesizing both global and local paths for mobile robots. DiPPeST achieves a $92\%$ success rate for simple environments and an average of $88\%$ success rate in generalization and edge case handling for environments up to 3.5 times harder than DiPPeR. We implement visual-servoing to allow real-robot execution of the novel local planning framework, achieving $80\%$ success rate in different environments and showcasing better behavior than two SOTA planners in a narrow passage experimental set-up. We identify that re-planning speed depends on inference time, which we aim to speed up by improving the denoising process. Additionally, goal-reaching is limited by the requirement of the global goal being visible within each frame, which we plan to address through the utilization of memory mechanism in the model or the use of Visual Odometry to better track features beyond a single frame. Our next steps focus on incorporating the kinodynamic properties of the robot during the trajectory generation process to achieve feasible real-world plans, which will enable the planner to navigate more complex environments with greater accuracy and efficiency.

%\clearpage

% \section*{Acknowledgments}

\addtolength{\textheight}{0cm}   % This command serves to balance the column lengths

%%%%%%%%%%%%%%%%%%%%%%%%%%%%%%%%%%%%%%%%%%%%%%%%%%%%%%%%%%%%%%%%%%%%%%%%%%%%%%%%
\bibliographystyle{IEEEtran}
\bibliography{IEEEabrv, iros_2024}
\end{document}